# Independence Concepts for Convex Sets of Probabilities

**Luis M. De Campos and Serafín Moral**
Departamento de Ciencias de la Computación e I.A.
Universidad de Granada, 18071 - Granada - Spain
e-mails: lci@robinson.ugr.es,smc@robinson.ugr

## Abstract

In this paper we study different concepts of independence for convex sets of probabilities. There will be two basic ideas for independence. The first is irrelevance. Two variables are independent when a change on the knowledge about one variable does not affect the other. The second one is factorization. Two variables are independent when the joint convex set of probabilities can be decomposed on the product of marginal convex sets. In the case of the Theory of Probability, these two starting points give rise to the same definition. In the case of convex sets of probabilities, the resulting concepts will be strongly related, but they will not be equivalent. As application of the concept of independence, we shall consider the problem of building a global convex set from marginal convex sets of probabilities.

## 1 INTRODUCTION

Convex sets of probabilities have been used as a model for unknown or partially known probabilities (Cano et al. 1991, Dempster 1967, Levi 1985, Stirling and Morrel 1991, Walley 1991). The basic idea is that if for a variable we do not have the exact values of probabilities, we may have a convex set of probability distributions. From a behavioural point of view the use of convex sets of probabilities was justified by Walley (1991). According to this author what distinguishes this theory from the Bayesian one is that imprecision in probability and utility models is admitted. Strict bayesians assume that for each event there is some betting rate you consider fair: you are ready to bet on either side of the bet. This rate determines the exact value of your subjective probability of the event. Convex sets of probabilities arose by assuming that for each event there is a maximum rate at which you are prepared to bet on it (determining its lower probability) and a minimum rate (determining its upper probability).

We consider convex sets with a finite set of extreme probabilities. This makes possible the calculations with convex sets: we have to carry out the operations for the finite set of extreme points.

In probability theory perhaps the most important concept is the concept of independence. The knowledge of independence relationships among a set of variables gives rise to the decomposition of the global probability in more elementary parts. This factorization is fundamental to represent and to calculate with probability distributions involving a non trivial number of variables (Lauritzen and Spiegelhalter 1988, Pearl 1988, Shenoy and Shafer 1990).

There are two main approaches to define independence:

- Irrelevance condition.- Two variables are independent if no piece of information about one of them can change our state of knowledge about the other.
- Decomposition condition.- Two variables are independent if the global information about the two variables can be expressed as a combination of two pieces of knowledge, one for each variable.

Both approachess are equivalent in the case of Classical Probability Theory, but things are not so easy for convex sets of probabilities. First, these conditions can have different interpretations leading to different definitions. The irrelevance property depends on the definition of conditioning that is being used and as it is well known there are different ways of doing conditioning in upper and lower probabilities (Moral and Campos 1991, Dubois and Prade 1994). Furthermore, the decomposition property can be applied to the individual probabilities or to the complete convex set of probabilities.

Several authors have considered different concepts of independence in the literature. Amarger et al. (1991) consider the decomposition property at the level of single probabilities. That is, independence can be expressed as a factorization of all the possible probabilities. This is called the sensitivity approach by Walley (1991). The decomposition property in terms of global convex sets has been considered by Shenoy (1994) and Cano et al. (1993). The irrelevance condition has been considered by Walley (1991), but only under one definition of conditioning, the so called upper and lower probabilities conditioning (Moral and Campos 1991, Dubois and Prade 1994). Campos and Huete (1993) have considered the definition of independence by means of the irrelevance condition with several definitions of conditioning, but they consider the model of upper and lower envelopes, a model which is more restrictive than general convex sets of probabilities.

The objective of this paper is to make an extensive study of independence in convex sets of probabilities. Section 2 introduces the essential concepts for convex sets of probabilities. Section 3 studies the differenct definitions of independence and their relationships. Section 4 is devoted to conditional independence. Finally section 5 considers the problem of building a global convex set from marginal convex sets. This is a problem strongly related with independence, because the hypothesis of independence usually allows to determine an only global representation of uncertainty with the given marginals.

## 2 CALCULUS WITH CONVEX SETS OF PROBABILITIES

In this section we describe a model for the calculation with convex sets of probabilities. Assume that we have a population $\Omega$ and a variable $X$ defined on $\Omega$ and taking its values on a finite set $U = \{u_1, ..., u_n\}$.

We shall consider that our knowledge about how $X$ takes its values is represented by a convex set of probabilities, $H^X$, with a finite set of extreme points $\text{Ext}(H^X) = \{p_1, ..., p_k\}$. Each $p_i$ is a probability distribution on $U$ and $\text{Ext}(H^X)$ are the extreme points of $H^X$.

Before going on we need to fix some notation. Assume that $h$ is a function from $U \times V$ onto $\mathbb{R}$ and $h'$ a function from $V \times W$ onto $\mathbb{R}$, then the multiplication of these two functions is a function, $h.h'$, defined on $U \times V \times W$ and given by, $h.h'(u,v,w) = h(u,v).h'(v,w)$.

We will interpret this definition on an extensive way. For example it will be applied also to the case in which $h'$ is defined on $V$ instead of $V \times W$. Then, $h.h'$ will be defined on $U \times V$ and we only have to drop coordinate $w$ in above expression: $h.h'(u,v) = h(u,v).h'(v)$. Analogously if $h$ is defined on $U$ and $h'$ on $W$, then $h.h'$ is defined on $U \times W$ with $h.h'(u,w) = h(u).h'(w)$.

This operation is extended to convex sets of functions. If $H$ and $H'$ are convex sets with $\text{Ext}(H) = \{h_1, ..., h_k\}, \text{Ext}(H') = \{h'_1, ..., h'_l\}$. Then the combination of $H$ and $H'$ will be the convex set, $H \otimes H'$ given by

$$H \otimes H' = \text{CH}\{h_1.h'_1, .., h_1.h'_l, ..., h_k.h'_1, .., h_k.h'_l\}$$

where CH stands for the convex hull operator (the minimum convex set containing a given set).

If $h$ is a function from $U \times V$ onto $\mathbb{R}$, then the marginal of $h$ to $U$ is the function $h^{\downarrow U}$ defined on $U$ and given by, $h^{\downarrow U}(u) = \sum_{v \in V} h(u,v)$.

The marginalization on $V$ is defined on an analogous way. This definition can be extended also to convex sets. If $H$ is a convex set of functions on $U \times V$, with extreme points, $\text{Ext}(H) = \{h_1, ..., h_k\}$, then the marginalization of $H$ to $U$ is the convex set given by, $H^{\downarrow U} = \text{CH}\{h_1^{\downarrow U}, ..., h_k^{\downarrow U}\}$

In the following, we give the elementary concepts to work with several variables under convex sets of probabilities. We shall assume that $X$ and $Y$ are variables taking values on finite sets $U$ and $V$, respectively.

If we have a global convex set $H^{X,Y}$ with extreme points $\{p_1, ..., p_k\}$ of possible bidimensional probabilities for variables $(X, Y)$, then we define the marginal convex sets, $H^X$ and $H^Y$, for variables $X$ and $Y$, as follows:

$$H^X = (H^{X,Y})^{\downarrow U} = \text{CH}\{p_1^{\downarrow U}, ..., p_k^{\downarrow U}\}$$

$$H^Y = (H^{X,Y})^{\downarrow V} = \text{CH}\{p_1^{\downarrow V}, ..., p_k^{\downarrow V}\}$$

From a global convex set $H^{X,Y}$ we can obtain also a conditional set, $H^{Y|X}$, given by

$$H^{Y|X} = \text{CH}\{p_1/p_1^{\downarrow U}, ..., p_k/p_k^{\downarrow U}\}$$

where the division stands for pointwise division, $p_i/p_i^{\downarrow U}(u,v) = p_i(u,v)/p_i^{\downarrow U}(u)$, being $0/0 = 0$.

If we start with some global set $H^{X,Y}$ and we calculate the marginal set $H^X$ and the conditional set $H^{Y|X}$, then the initial global set can not always be recovered from $H^X$ and $H^{Y|X}$. In effect, if we calculate the combination of these two sets, we will obtain the global set $H'^{X,Y} = H^X \otimes H^{Y|X}$. However, in general, we will have that $H^{X,Y}$ is included into $H'^{X,Y}$, but they will not be always equal. This situation is different of the case of probability theory. When we have a single probability distribution, the global probability can be obtained from a marginal and a conditional, but this is not true for convex sets. $H'^{X,Y}$ is the biggest convex set having $H^X$ as marginal and $H^{Y|X}$ as conditional. It is calculated by multiplying each point in $H^X$ by each point in $H^{Y|X}$ and then taking the convex hull. The elements of $H^X$ are the marginal probabilities of the probabilities in $H^{X,Y}$. Analogously, the elements of $H^{Y|X}$ are the conditional probabilities. The difference with $H'^{X,Y}$ is that in $H^{X,Y}$ we do not have necessarily the combination of all the marginal functions on $H^X$ and all the conditional functions in $H^{Y|X}$. What we know is that for every marginal in $H^X$ there is at least one conditional in $H^{Y|X}$, such that their product is in $H^{X,Y}$ and that for every conditional in $H^{Y|X}$ there is at least one marginal in $H^X$, with the product of the two in $H^{X,Y}$.

This problem has some relationship with the determination of causal relationships between variables. In general, when $X$ is a cause of $Y$, the determination of the marginal probability in $X$ and the conditional probability of $Y$ given $X$ should be independent (Spirtes et al. 1993). In such a case, if $H^{X,Y}$ is a global convex set for $X$ and $Y$, we should have $H^{X,Y} = H^X \otimes H^{Y|X}$. Taking this idea as basis, but without pretending to characterize causal relationships, we will say that $X$ is a cause of $Y$ under $H^{X,Y}$ if and only if $H^{X,Y} = H^X \otimes H^{Y|X}$.

Now, we consider the problem of conditioning in the sense of focusing (Dubois and Prade 1994), that is when we incorporate observations for a particular case to general probabilistic knowledge. First we shall consider the definition of conditioning proposed by Moral and Campos (1991).

Assume a convex set for variable $X$: $H^X = \mathrm{CH}\{p_1,\ldots,p_k\}$ and that we have observed '$X$ belongs to $A$', then the result of conditioning is the convex set, $H^X|_1 A$, generated by points $\{p_1.l_A,\ldots,p_k.l_A\}$ where $l_A$ is the likelihood associated with set $A$ ($l_A(u) = 1$, if $u \in A; l_A(u) = 0$, otherwise). That is $H^X|_1 A = H^X \otimes \{l_A\}$

It is important to remark that $H^X|_1 A$ is a convex set of differently normalized functions. If we call $r = \sum_{u \in U} p(u).l_A(u) = p(A)$, then by calculating $(p.l_A)/r$ (when $r \neq 0$) we get the conditional probability distribution $p(.|A)$. The set $H^X|_2 A = \{p(.|A) : p \in H, p(A) \neq 0\}$ was proposesed by Dempster (1967) as the set of conditioning, and has been widely used. However by considering only this set, we loose information. The reason being that, by normalizing each probability, we forget the normalization values, $r = p(A)$, which are a likelihood induced by the observation on the set of possible probability distributions.

If $H_1$ and $H_2$ are two convex sets of non-necessarily normalized functions we will consider that they are equivalent if and only if there is an $\alpha > 0$ such that $\mathrm{CH}(H_1 \cup \{h_0\}) = \mathrm{CH}(H_2 \cup \{h_0\}).\alpha$, where $h_0$ is the null function: $h_0(u) = 0, \forall u \in U$. Reasons for this equivalence are given by Cano et al. (1991). The underlying idea is that multiplying all the functions of the convex set by the same real number we get an equivalent set. It says also that the presence of the null function does not change our state of belief.

These two definitions can be extended to the case in which $l$ is a general likelihood function, $l : U \to [0,1]$. $H^X|_1 l$ is equal to $H^X \otimes \{l\}$ and $H^X|_2 l$ is defined as $\{p.l/r : p \in H^X, r = \sum_{u \in U} p.l(u), r \neq 0\}$.

If we have variables $X$ and $Y$ taking values on $U$ and $V$ respectively and $H^{X,Y}$ is a global convex set of probabilities for these two variables, then by $H^X|_1(Y \in B)$ we will denote the Moral and Campos conditioning of $H^{X,Y}$ to the set $U \times B$ and the marginalization of the result to $U$. That is, $H^X|_1(Y \in B) = (H^{X,Y}|_1 U \times B)^{\downarrow U}$. Analogously, for the Dempster conditioning we will consider $H^X|_2(Y \in B) = (H^{X,Y}|_2 U \times B)^{\downarrow U}$.

If $B = \{v\}$, then $H^X|_i(Y \in B)$ will be denoted as $H^X|_i(Y = v)$ ($i = 1, 2$).

If $l_Y$ is a likelihood function about $Y$, $l_Y : V \to [0,1]$, then $H^X|_1 l_Y$, will denote the Moral and Campos conditioning of $H^{X,Y}$ to the likelihood on $U \times V$ given by $l(u,v) = l_Y(v)$ and the marginalization of the result to $U$. That is $H^X|_1 l_Y = (H^{X,Y} \otimes \{l_Y\})^{\downarrow U}$. On the same way, $H^X|_2 l_Y$ will denote the Dempster conditioning of $H^{X,Y}$ to $l$ and the posterior marginalization to $U$.

## 3 INDEPENDENCE

Assume that we have a two-dimensional variable $(X,Y)$ taking values on the cartesian product $U \times V$. In this section we shall consider the conditions under which variables $X$ and $Y$ can be considered as independent, when the global information about these variables is given by a convex set of probabilities. Previously we will recall the definition of independence for a single probability distribution.

**Definition 1** *We say that $X$ and $Y$ are independent under global probability $p$, if and only if one of the following equivalent conditions is verified*

1. $p(u,v) = p^{\downarrow U}(u).p^{\downarrow V}(v), \forall (u,v) \in U \times V$
2. $p(u|v) = p^{\downarrow U}(u), \forall (u,v)$ with $p^{\downarrow V}(v) > 0$

*where $p(u|v) = p(u,v)/p^{\downarrow V}(v)$ is the conditional probability.*

A first definition of independence when we have more than an only probability distribution is to assume that all the possible probabilities verify above condition. However this condition is too strong if we want to work with convex sets, as it is shown by the following theorem.

**Theorem 1** *Assume that $H^{X,Y}$ is a non-empty convex set of probabilities defined on $U \times V$, then if*

$$H^{X,Y} \subseteq \{p.p' : p \in H^X, p' \in H^Y\}$$

*we have that some of the marginal sets $H^X$ or $H^Y$ is trivial in the sense that it contains only one probability distribution.*

To extend this definition of independence on a non-trivial way to the case of imprecise probabilities, we propose two alternatives: the first is to define independence for general sets (without assuming that the knowledge is always represented by means of a convex set); the second is to impose the condition of independence only for the extreme points of the convex set.

**Definition 2 (Type-1)** *If $H$ is a set of joint probability distributions (non necessarily convex) for $(X, Y)$, then we say that $X$ and $Y$ are type-1 independent, which will be denoted as $I_1(X, Y)$, if and only if for every $p \in H$ it is verified: $p(u, v) = p^{\downarrow U}(u).p^{\downarrow V}(v), \forall (u, v) \in U \times V$*

**Definition 3 (Type-2)** *If $H^{X,Y}$ is a convex set of probability distributions for $(X, Y)$, then we say that $X$ and $Y$ are type-2 independent, which will be denoted as $I_2(X, Y)$, if and only if for every $p \in Ext(H^{X,Y})$ it is verified: $p(u, v) = p^{\downarrow U}(u).p^{\downarrow V}(v), \forall (u, v) \in U \times V$*

Type-1 independence is called by Walley (1991) the sensitivity analysis approach to independence. Although these definitions look as a very natural extension of independence they have severe inconvenients as the following example shows.

**Example.-** Consider $U = \{u_1, u_2\}, V = \{v_1, v_2\}$ and the probabilities $p_1$ and $p_2$, on $U \times V$ given by

$$p_1(u_1, v_1) = 1, \quad p_1(u_i, v_j) = 0, \text{ otherwise}$$
$$p_2(u_2, v_2) = 1, \quad p_2(u_i, v_j) = 0, \text{ otherwise}$$

If $H = \{p_1, p_2\}$, we obtain that $X$ and $Y$ are type-1 independent. However, this is not very intuitive: under $H$ there is functional dependence between $X$ and $Y$. The only possible pairs are $(u_1, v_1)$ and $(u_2, v_2)$. Therefore, if we know that $X = u_1$ then we obtain that $Y = v_1$ and if $X = u_2$ we have $Y = v_2$.

Things are very similar if we assume that we have the convex set generated by these two probabilities: CH($H$). We have type-2 independence and the same functional dependence between the variables  ∎

The problem with above example is that each single probability distribution determines an independent relationship between the variables, but the two probabilities at the same time determine a functional relationship. In general, independence is very related with decomposition. But the appropriate way of defining independence in terms of decomposition is not by assuming that every probability is decomposable, but by assuming that the global set is decomposable. This idea was proposed by Shenoy (1994) for the case of general abstract valuations and it is applied to the particular case of convex set of probabilities in the following definition.

**Definition 4 (Type-3)** *If $H^{X,Y}$ is a global convex set of probabilities for $(X, Y)$, we say that $X$ and $Y$ are type-3 independent, $I_3(X, Y)$, if and only if,*

$$H^{X,Y} = H^X \otimes H^Y$$

*where $H^X$ and $H^Y$ are the marginal sets of $H^{X,Y}$ on $U$ and $V$ respectively*

In the case of the example it is clear that there is not type-3 independence. It is also immediate to show the following proposition.

**Proposition 1** *If $I_3(X, Y)$, then $I_2(X, Y)$.*

One of the most intuitive ways of defining independence is by means of the concept of irrelevance (Walley 1991, Campos and Huete 1993): if nothing that can be learnt about $Y$ can change our state of knowledge about $X$. The way of introducing an observation about $Y$ is by means of conditioning. As there are two main ways of conditioning in convex sets of probabilities, then taking as basis this idea, we get two concepts of independence.

**Definition 5 (Type-4)** *We say that variable $X$ is type-4 independent of variable $Y$, $I_4(X, Y)$, under convex set $H^{X,Y}$ if and only if for every likelihood $l_Y$ defined on $V$, we have that $H^X|_1 l_Y$ is equivalent to $H^X$ or to $\emptyset$.*

**Definition 6 (Type-5)** *We say that variable $X$ is type-5 independent of variable $Y$, $I_5(X, Y)$, under convex set $H^{X,Y}$ if and only if for every likelihood $l_Y$ defined on $V$, we have that $H^X|_2 l_Y = H^X$ or $H^X|_2 l_Y = \emptyset$.*

The following proposition is immediate from the definitions of type-4, type-5 independence and conditioning.

**Proposition 2** *If $I_4(X, Y)$ then $I_5(X, Y)$.*

On the contrary to the case of a single probability, for type-4 and type-5 independence the equality of the marginal and the conditional information to the events $[Y = v]$ is not sufficient to assure the equality of the marginal and the conditional to an arbitrary likelihood on $Y$: $l_Y$.

**Example.-** Assume that $U = \{u_1, u_2\}, V = \{v_1, v_2, v_3\}$ and the convex set $H^{X,Y}$ with the extreme points, $p_1, p_2, p_3, p_4$, given by:

|   | $(u_1, v_1)$ | $(u_1, v_2)$ | $(u_1, v_3)$ | $(u_2, v_1)$ | $(u_2, v_2)$ | $(u_2, v_3)$ |
|---|---|---|---|---|---|---|
| $p_1$ | 1/3 | 0 | 0 | 2/3 | 0 | 0 |
| $p_2$ | 1/4 | 0 | 0 | 3/4 | 0 | 0 |
| $p_3$ | 0 | 0.1 | 0.2 | 0 | 0.3 | 0.4 |
| $p_4$ | 0 | 0.15 | 2/15 | 0 | 0.45 | 4/15 |

In this case, it is easy to show that $\forall v \in V, H^X = H^X|_2(Y = v)$. However, if we calculate the conditional information to the likelihood, $l_Y$, given by

$$l_Y(v_1) = 0, l_Y(v_2) = 1, l_Y(v_3) = 1$$

then we obtain a different convex set.

For the case of type-4 independence, above example is valid too. All the conditional sets $H^X|_1(Y = v)$ are equivalent to $H^X$, but the conditional set to likelihood $l_Y$ gives rise to a non-equivalent convex set. ∎

Type-4 and Type-5 independences do not imply type-2 independence as the following example shows.

**Example.-** Consider $U = \{u_1, u_2\}, V = \{v_1, v_2\}$, and the convex set $H^{X,Y}$ with extreme points, $p_1, p_2, p_3$, given by

|   | $(u_1, v_1)$ | $(u_1, v_2)$ | $(u_2, v_1)$ | $(u_2, v_2)$ |
|---|---|---|---|---|
| $p_1$ | 0.24 | 0.56 | 0.06 | 0.14 |
| $p_2$ | 0.15 | 0.35 | 0.15 | 0.35 |
| $p_3$ | 0.15 | 0.56 | 0.15 | 0.14 |

In these conditions, we have $I_4(X, Y)$ and $I_5(X, Y)$, but $p_3$ is an extreme point which is not decomposable as product of its marginal probability distributions.

This example shows also that type-4 and type-5 independences are not symmetrical. In fact, we have that $X$ is type-4 (and type-5) independent of $Y$, but $Y$ is neither type-4 nor type-5 independent of $X$. ∎

There is a relationship between type-4, type-5 independences and type-2 independence. In fact, in both cases we can prove the following theorem, which says that although not all the single extreme probabilities show independence between $X$ and $Y$ there are necessarily some extreme probabilities under which there is $X,Y$ independence.

**Theorem 2**    - If $I_4(X,Y)$ then $H^X \otimes H^Y \subseteq H^{X,Y}$
- If $I_5(X,Y)$ then $\forall p \in Ext\,(H^Y), \exists q \in H^X$, such that $p.q \in H^{X,Y}$

Under type-4 independence, we do not have necessarily the equality $H^X \otimes H^Y = H^{X,Y}$, as was shown in above example. To get the decomposition of type-3 independence, we should add some additional condition to type-4 independence. The combination of type-4 and type-2 independence is appropriate for this purpose.

**Theorem 3** $I_3(X,Y)$ if and only if $I_2(X,Y)$ and $I_4(X,Y)$.

Under type-2 independence type-4 and type-5 independence are symmetrical concepts and we do not need to specify which variable is independent of the other.

We have determined another properties which are always verified under type-3 independence and not under type-4 independence, as the following one: for all convex set of probabilities for variables $X, Y$, and $Z$, such that $X, Y$ has $H^{X,Y}$ as marginal information and such that $X, Y$ is a cause of $Z$, we have that $X$ is a cause of $Z$ and $Y$ is a cause of $Z$. In other words, for every convex set about the three variables $H^{X,Y,Z}$, if this set can be decomposed as $H^{X,Y} \otimes H^{Z|X,Y}$, then the marginal on $X, Z$, $H^{X,Z}$ can be decomposed as $H^X \otimes H^{Z|X}$ and the marginal on $Y, Z$, decomposed as, $H^{Y,Z} = H^Y \otimes H^{Z|Y}$. This is a very natural property under independence of $X$ and $Y$, which is verified under type-3 independence, but not under type-4 independence. However, we have not proved yet, whether this property together with type-4 independence is equivalent to type-3 independence.

There is also an strong relationship between type-2 and type-5 independence.

**Theorem 4** If $I_2(X,Y)$, then $I_5(X,Y)$ if and only if $\forall p \in H^X, \forall v \in V$, if $\exists p' \in H^Y$ with $p'(v) > 0$, we have that $\exists p'' \in H^Y$ with $p''(v) > 0$ and $p.p'' \in H^{X,Y}$

Most of the differences among the different notions of independence come from the correlations which can exist between marginal and conditional probabilities in a global convex set $H^{X,Y}$. In fact, if $X$ is a cause of $Y$ (or viceversa) all the different independencies for convex sets are equivalent.

**Theorem 5** If $X$ is a cause of $Y$ under $H^{X,Y}$, that is $H^{X,Y} = H^X \otimes H^{Y|X}$, then $I_2(X,Y) \Leftrightarrow I_3(X,Y) \Leftrightarrow I_4(X,Y) \Leftrightarrow I_5(X,Y)$

## 4  CONDITIONAL INDEPENDENCE

The concept of independence reaches a big degree of expressiveness when one can talk about conditional independence. $X$ and $Y$ are conditionally independent given $Z$, when they are independent under a perfect knowledge of the value of $Z$. First we give a precise meaning of this concept in the case of classical probability theory. It will be assumed that $X, Y$, and $Z$ are three variables taking values on $U, V$, and $W$, respectively.

**Definition 7** *We say that variables $X$ and $Y$ are conditionally independent given $Z$ under global probability $p$, if and only if one of the following equivalent conditions is verified*

1. $p(u,v|w) = p^{\downarrow U \times W}(u|w) \cdot p^{\downarrow V \times W}(v|w)$,
   $\forall (u,v,w)$ with $p^{\downarrow W}(w) > 0$
2. $p(u,v,w) = p^{\downarrow V \times W}(v,w) \cdot p^{\downarrow U \times W}(u|w)$,
   $\forall (u,v,w)$ with $p^{\downarrow W}(w) > 0$

For the case of imprecise probabilities, the definitions of marginal independence can be extended to the case of conditional independence.

**Definition 8 (Type-1)** *If $H$ is a set of joint probability distributions (non necessarily convex) for $(X,Y,Z)$, then we say that $X$ and $Y$ are type-1 conditionally independent given $Z$, which will be denoted as $I_1(X,Y|Z)$, if and only if Definition 7 is verified for every $p \in H$.*

**Definition 9 (Type-2)** *If $H^{X,Y,Z}$ is a convex set of probability distributions for $(X,Y,Z)$, then we say that $X$ and $Y$ are type-2 conditionally independent given $Z$, $I_2(X,Y|Z)$, if and only if Definition 7 is verified for every $p \in Ext(H^{X,Y,Z})$.*

**Definition 10 (Type-3)** *If $H^{X,Y,Z}$ is a global convex set of probabilities for $(X,Y,Z)$, we say that $X$ and $Y$ are type-3 conditionally independent given $Z$, $I_3(X,Y|Z)$, if and only if,*

$$H^{X,Y,Z} = H_1 \otimes H_2$$

*where $H_1$ is a convex set of functions on $U \times W$ and $H_2$ is a convex set of functions on $V \times W$.*

Let us remark that in above definition there is no necessity that $H_1$ or $H_2$ is a marginal or conditional convex set obtained from $H^{X,Y,Z}$. In fact in some occasions only it is only possible a decompostion in which $H_1$ contains the conditional probabilities of $X$ given $Z$ and a *part* of the marginal information about $Z$. To obtain a marginal distribution about $Z$, we have to multiply this part with the one in the other convex set $H_2$. That was not the case of unconditional independence. If a bidimensional convex $H^{X,Y}$, is decomposed as product of a convex set on $U$ and a convex set about $V$, we can always assume that these convex sets are the marginal sets of $H^{X,Y}$.

**Definition 11 (Type-4)** *We say that variable $X$ is type-4 conditionally independent of variable $Y$ given $Z$, $I_4(X,Y|Z)$, under convex set $H^{X,Y,Z}$ if and only if for every likelihood $l_Y$ defined on $V$ and every $w \in W$ we have that $H^X|_1(l_Y, Z = w)$ is equivalent to $H^X|_1(Z = w)$ or to $\emptyset$.*

**Definition 12 (Type-5)** *We say that variable $X$ is type-5 conditional independent of variable $Y$ given $Z$, $I_5(X,Y|Z)$, under convex set $H^{X,Y,Z}$ if and only if for every likelihood $l_Y$ defined on $V$ and every $w \in W$, we have that $H^X|_2(l_Y, Z = w) = H^X|_2(Z = w)$ or $H^X|_2(l_Y, Z = w) = \emptyset$.*

**Theorem 6** *If $H^{X,Y,Z}$ is a global convex set for variables $X,Y,Z$, then we have the following implications.*

$$I_3(X,Y|Z) \Longrightarrow I_2(X,Y|Z), I_4(X,Y|Z)$$
$$I_4(X,Y|Z) \Longrightarrow I_5(X,Y|Z)$$

On the contrary to the case of unconditional independence, we can have type-2 and type-4 conditional independence without having type-3 independence.

**Example.-** Assume that $W = \{0,1\}$ and that $H^{X,Y,Z}$ is the convex hull generated by points: $p_1(u,v,w) = p_1(u|w).p(w).p'_1(v|w), p_2(u,v,w) = p_2(u|w).p(w).p'_2(v|w)$, where for $w = 0$ we have that $p_1(.|0) = p_2(.|0), p'_1(.|0) \neq p'_2(.|0)$ and for $w = 1$ we have $p'_1(.|1) = p'_2(.|1), p_1(.|1) \neq p_2(.|1)$.

In these conditions, there is type-2 and type-4 conditional independence but not type-3 independence. ∎

We can obtain interesting results when we mix conditional independence with causal relationships.

**Theorem 7** *If $H^{X,Y,Z}$ is a global convex set, such that one of the following conditions is verified:*

*a) $Z$ is a cause of $X,Y$*

*b) $X,Z$ is a cause of $Y$*

*c) $Y,Z$ is a cause of $X$*

*then $I_2(X,Y|Z) \Leftrightarrow I_3(X,Y|Z) \Leftrightarrow I_4(X,Y|Z) \Leftrightarrow I_5(X,Y|Z)$.*

*Furthermore if there is conditional independence, we have the following decompositions in each one of the cases:*

*a) $H^{X,Y,Z} = H^Z \otimes H^{X|Z} \otimes H^{Y|Z}$*

*b) $H^{X,Y,Z} = H^{X,Z} \otimes H^{Y|Z}$*

*c) $H^{X,Y,Z} = H^{Y,Z} \otimes H^{X|Z}$*

## 5 THE MARGINAL PROBLEM

In this section and as an application of the concept of independence, we consider the following problem: assume that we have three variables, $X, Y$, and $Z$ and that we have two convex sets of probabilities, $H_1$, about variables $X, Z$ and $H_2$, about variables $Y, Z$. The question is how to build on a reasonable way a global set $H^{X,Y,Z}$ from this marginal information.

For the case of a single probability distribution, if $p_1$ is a marginal probability for variables $X, Z$ and $p_2$ is a marginal probability for variables $Y, Z$, then the problem is solved in two steps:
1. We check the compatibility of the two marginal distributions. That is, whether $p_1^{\downarrow W} = p_2^{\downarrow W}$.
2. In the case of compatibility, we fix a criterion allowing us to determine an unique probability among all those $p$ on $U \times V \times W$ having $p_1$ and $p_2$ as marginals. The most usual hypothesis is to assume that $X$ and $Y$ are conditional independent given $Z$. Then, we obtain the probability:

$$p(u, v, w) = p_1(u, w) \cdot p_2(v, w) / p_1^{\downarrow W}(w)$$

For convex sets of probabilities, the steps are:
1. Make $H_1$ and $H_2$ consistent, by transforming them into $H_1'$ and $H_2'$ given by,

$$H_1' = \{p \in H_1 : p^{\downarrow W} \in H_2^{\downarrow W}\}$$

$$H_2' = \{p \in H_2 : p^{\downarrow W} \in H_1^{\downarrow W}\}$$

If $H_1' = H_2' = \emptyset$ then we have inconsistency. In other case, $H_1'^{\downarrow W} = H_2'^{\downarrow W} \neq \emptyset$ and we have been able of reaching consistency.

2. In the case of consistency, we have to assume a condition on $X, Y,$ and $Z$, allowing to determine an unique convex set of probabilities, $H$, defined on $U \times V \times W$, with $H_1'$ and $H_2'$ as marginals. This problem is not as easy as in the case of a single probability distribution. An initial difficulty is clear: we have several definitions of independence. The rest of the section is devoted to give a reasonable answer to this question.

We consider that $H_1$ and $H_2$ are two compatible convex sets, that is, $H_1^{\downarrow W} = H_2^{\downarrow W} \neq \emptyset$. To determine a global convex set under an assumption of independence, the firsts thing that we can observe is that not always we can assume type-3 conditional independence. Type-3 independence is the strongest condition we have studied and it is a first candidate to be considered as hypothesis to fix a global $H$. The reason is that this definition is directly related with the idea of decomposition. However, the following example shows that we can have two compatible $H_1$ and $H_2$ for which there is no global set $H$ having $H_1$ and $H_2$ as marginals with type-3 conditional independence.

**Example.-** Assume $U = V = W = \{0, 1\}$. Let $H_1$ be the convex set on $U \times W$ given by extreme points:

$p_1^1(0,1) = 0.99, p_1^1(1,0) = 0.01, p_1^1(u,w) = 0,$ otherwise
$p_2^1(1,1) = 1, p_2^1(u,w) = 0,$ otherwise

And $H_2$ the convex set on $V \times W$ given by:

$p_1^2(0,1) = 0.99, p_1^2(1,0) = 0.01, p_1^2(v,w) = 0,$ otherwise
$p_2^2(1,1) = 1, p_2^2(v,w) = 0,$ otherwise

In this case, there is an only convex set with type-2 independence and with $H_1$ and $H_2$ as marginal sets, the one given by the extreme probabilities $p_1$ and $p_2$ on $U \times V \times W$ defined by:

$$p_i(u, v, w) = p_i^1(u, w) \cdot p_i^2(v|w), \quad i = 1, 2$$

However with this convex set we do not have type-3 independence.    ∎

Given the definitions, there seems more difficult to work with type-4 and type-5 independence. So the natural candidate is to impose type-2 independence as hypothesis. There is another difficulty with this hypothesis. There is not always an unique convex set compatible with given marginals and verifying type-2 independence. As example, consider that $H_1 = \text{CH}(\{p_1^1, p_2^1\})$ and $H_2 = \text{CH}(\{p_1^2, p_2^2\})$ in such a way that $p_1^{1\downarrow W} = p_2^{1\downarrow W} = p_1^{2\downarrow W} = p_2^{2\downarrow W}$. In this case we have compatibility between $H_1$ and $H_2$, but there are several convex sets having $H_1$ and $H_2$ as marginal sets and verifying type-2 conditional independence of $X$ with respect to $Y$ given $Z$. Among them, we could consider the convex set generated by points: $p_1^1 \cdot p_1^2/p_1^{2\downarrow W}, p_2^1 \cdot p_2^2/p_2^{2\downarrow W}$; or the convex set generated by points: $p_1^1 \cdot p_2^2/p_2^{2\downarrow W}, p_2^1 \cdot p_1^2/p_1^{2\downarrow W}$

However, there is always a least specific convex set verifying type-2 conditional independence (in the sense that it contains every other convex set verifying this hypothesis). In this case, this convex set is given by,

$$H^{X,Y,Z} = \text{CH}\left(\left\{\frac{p_1^1 \cdot p_1^2}{p_1^{2\downarrow W}}, \frac{p_2^1 \cdot p_2^2}{p_2^{2\downarrow W}}, \frac{p_1^1 \cdot p_2^2}{p_2^{2\downarrow W}}, \frac{p_2^1 \cdot p_1^2}{p_1^{2\downarrow W}}\right\}\right)$$

In general we can prove the following theorem.

**Theorem 8** *If $H_1$ is a convex set of probabilities on $U \times W$ and $H_2$ is a convex set of probabilities on $V \times W$, such that $H_1^{\downarrow W} = H_2^{\downarrow W}$, then the convex set, $H^{X,Y,Z}$, on $U \times V \times W$, given by*

$$H^{X,Y,Z} = CH\left(\left\{\frac{p_1 \cdot p_2}{p_2^{\downarrow W}} : p_1 \in H_1, p_2 \in H_2, p_1^{\downarrow W} = p_2^{\downarrow W}\right\}\right)$$

*verifies the following properties,*

1. $(H^{X,Y,Z})^{\downarrow U \times W} = H_1, (H^{X,Y,Z})^{\downarrow V \times W} = H_2$
2. $I_2(X, Y|Z)$ under $H^{X,Y,Z}$
3. *Every convex set of probabilities $H$ on $U \times V \times W$ verifying 1 and 2, is included in $H^{X,Y,Z}$*

This will be our proposal for the construction of a global convex set from compatible marginal sets, $H_1$ and $H_2$. The global convex set, $H^{X,Y,Z}$ built in above theorem will be denoted as $H_1 \odot H_2$. There is another result justifying this selection: if there is a global set under which there is type-3 conditional independence then $H^{X,Y,Z}$ will coincide with this set. This is stated in the following theorem.

**Theorem 9** *If If $H_1$ is a convex set of probabilities on $U \times W$ and $H_2$ is a convex set of probabilities on $V \times W$, such that $H_1^{\downarrow W} = H_2^{\downarrow W}$, and $H$ is a global convex set on $U \times V \times W$, verifying:*

1. $H^{\downarrow U \times W} = H_1, H^{\downarrow V \times W} = H_2$
2. $I_3(X, Y|Z)$ under $H$.

*Then $H = H^{X,Y,Z} = H_1 \odot H_2$.*

## 6 CONCLUSIONS

In this paper we have studied different concepts of independency for convex sets of probabilities. The situation here is different than in the case of a single probability distribution. Several generalizations of the concept of independence are possible. There is one which is the strongest of all the definitions we have proposed: type-3 independence. Our feeling is that this is the most appropriate definition of independence and that the other definitions consider only partial aspects of the concept of independence. However, some of then can be useful in other situations as the case of type-2 independence in the marginal problem.

The main differences among the different definitions come from the correlation that may exists among the probabilities governing the variables. For example, in type-2 independence if we fix a probability, we have independence, but the problem is that the determination of the probability can establish a strong relationship between the values of the variables.

Some aspects of the concept of independence have not been studied by the space limitations of this paper. Among others to check whether graphoid axioms are verified and to study other applications of the concept of independence as the learning of causal networks from data (Pearl 1988). Of special interest is for us the study of the relationships between decomposition properties of convex sets and the determination of causal relationships.

### Acknowledgements

This work has been supported by the DGICYT under Project n. PB92-0939.